\documentclass[10pt]{cai}
\usepackage{booktabs}
\usepackage{graphicx}

\usepackage[font=small,skip=4pt]{caption}

\graphicspath{{figs/}}

\usepackage{lmodern}
\usepackage{csquotes}

\usepackage[english]{babel}  
\addto\extrasenglish{
  
}
\usepackage{hyperref}

\newcommand{\guillemet}[1]{``#1''}

\definecolor{green}{HTML}{82B366}
\definecolor{blue}{HTML}{6C8EBF}
\definecolor{red}{HTML}{B85450}
\definecolor{gray}{HTML}{666666}

\begin{document}

\volumeheader{36}{0}
\begin{center}

  \title{RISC: Generating Realistic Synthetic Bilingual Insurance Contract}
  \maketitle

  \thispagestyle{empty}
  \begin{tabular}{cc}
    David Beauchemin\upstairs{\affilone,*}, Richard Khoury\upstairs{\affilone}
   \\[0.25ex]
   {\small \upstairs{\affilone} Université Laval} \\
  \end{tabular}
  
  \emails{
    \upstairs{*}david.beauchemin@ift.ulaval.ca 
    }
  \vspace*{0.2in}
\end{center}

\begin{abstract}
This paper presents RISC, an open-source Python package data generator\footnote{\href{https://github.com/GRAAL-Research/risc}{https://github.com/GRAAL-Research/risc}}. RISC generates look-alike automobile insurance contracts based on the Quebec regulatory insurance form in French and English. Insurance contracts are 90 to 100 pages long and use complex legal and insurance-specific vocabulary for a layperson. Hence, they are a much more complex class of documents than those in traditional NLP corpora. Therefore, we introduce RISCBAC, a Realistic Insurance Synthetic Bilingual Automobile Contract dataset based on the mandatory Quebec car insurance contract. The dataset comprises 10,000 French and English unannotated insurance contracts. RISCBAC enables NLP research for unsupervised automatic summarisation, question answering, text simplification, machine translation and more. Moreover, it can be further automatically annotated as a dataset for supervised tasks such as NER.
\end{abstract}

\copyrightnotice

\section{Introduction}
\label{sec:intro}
Application of NLP deep learning techniques on specialized domains 
have seen an increase in interest in recent years 
\cite{katz2023natural}.
The legal domain is one such domain, which is known to be complex and hermetic for a layperson \cite{beauchemin2020generating}. This complexity has real consequences for many individuals and organizations. For example, a Canadian study (in the province of Quebec) has shown that the public register of official court traces (i.e. dockets) of all legal cases lacks intelligibility to most citizens \cite{tep2019legal, parada2020digital}. Moreover, this complexity has raised concerns about assisting the public with fair access to justice and judicial information 
\cite{barton2017rebooting, susskind2019online}, especially after the COVID pandemic judicial system has taken overdue in their court cases \cite{rusakova2021challenges, matyas2021imagining}.

Even though judiciary systems produce, consume and use massive volumes of textual information \cite{chalkidis2021lexglue}, they lack technological solutions to increase their efficiency. Moreover, legal documents are known to be complex and lengthy and use specialized vocabulary \cite{katz2023natural}, which raises the technical challenge of developing NLP systems in that domain. 

Thus, creating curated large legal annotated corpora has been proven to be costly \cite{https://doi.org/10.48550/arxiv.2103.06268, wang2023maud}. For example, MAUD, an expert-annotated merger agreement understanding dataset, has been estimated to cost \$5 million using the standard hourly fees of specialized lawyers \cite{wang2023maud}. Despite the challenges, there has understandably been great interest in exploring the possibility of deep learning techniques such as the use of Transformer architecture (i.e. GPT-like model) \cite{bommarito2022gpt, choi2023chatgpt} for helping process complex legal texts.

Insurance contracts are a particular case of legal documents where documents are relatively standardized, yet they use legal and insurance-specific vocabulary. For example, they use long and wordy sentences to specify a property or life risk coverage. Also, insurance contracts (at least in Canada) use a base form that specifies many exclusions and limited coverage and use appended endorsements to modify the base form. Thus, the overall document, composed of a base form and endorsements, \guillemet{contredict} itself and must be interpreted as a whole.

Insurance products can represent significant financial implications for individual financial health in the event of a loss. For example, a residential property total loss represents a heavy loss for any individual. This situation has led many governments to establish insurance regulators such as the \textit{Autorité des marchés financiers} (AMF) in the province of Quebec \cite{loiamf}. Moreover, some insurance products are mandatory by law; for example, car civil liability insurance is mandatory in Quebec. Thus, choosing the right product is an essential step for many individuals, yet it is complicated. Regulations usually enforce a professional's advisory role as a legal obligation to insurers to protect the public \cite{loiamf}. However, in recent years, many governments have started authorizing the online sale of insurance products without the intervention of any human agent \cite{bill141, loiamf}. This new way of selling insurance has raised concerns for regulatory and professional organizations in their role to protect the public \cite{RCCAQ, memoireamfloi141}. It created an interest in leveraging new technologies, such as deep learning, to improve (or automate) access to more understandable and personalized information about insurance products. However, no insurance contract corpora are currently available to train machine learning (ML) models to tackle NLP tasks that apply to the insurance field \cite{katz2023natural}.

One of the particularities of insurance contracts is that they include detailed customer personal data such as name, date of birth and address. It is more challenging to release a public dataset based on actual customer insurance contracts since data would have to be anonymized. 
Moreover, they also include corporate property, namely the premium for a specific customer. Even if insurance contracts could be perfectly anonymized, releasing the premium could expose the insurer to premium reverse engineering from other insurers. For those reasons, in partnership with a Canadian insurance company, we have created a realistic insurance synthetic contract dataset generator based on our strong field expertise in the insurance domain and use as much real data as possible.

This paper's contributions are twofold: a realistic insurance synthetic contract data generator and a new synthetic automobile insurance contract dataset. It is outlined as the following, first, we study the available legal corpora and synthetic dataset generator in \autoref{sec:rel_work}. Then, we propose RISC, an open-source Python package, to generate realistic insurance synthetic contract datasets in \autoref{sec:risk}. Finally, in \autoref{sec:rsbaic}, we propose a realistic synthetic bilingual automobile insurance contract corpus based on Quebec's car insurance, and we discuss the ML research task enabled by this more difficult corpus as traditional NLP corpora.

\section{Related Work}
\label{sec:rel_work}

In recent years, a few legal corpora have been proposed in English, such as LEDGAR \cite{tuggener2020ledgar}, CUAD \cite{https://doi.org/10.48550/arxiv.2103.06268}, BillSum \cite{Eidelman_2019}, MAUD \cite{wang2023maud}, and EUR-Lex-Sum \cite{aumiller2022eur}. The first, LEDGAR, consists of 100,000 provisions to be classified as provisions types (e.g. law compliance). Provisions are the \guillemet{items} in any contract that constitute the contract's legal speech act. These provisions were extracted from contracts on the U.S. Securities and Exchange Commission (SEC) website, namely contracts between companies. The second, CUAD, is a dataset of 510 annotated contracts also used for classification, but for clause identification instead of provisions. However, these contracts are not insurance but rather reviews of general contracts to asses the rights or obligations of an individual or company. The third, BillSum, consists of 22,218 US Congressional bills and reference summaries for legal text summarization. The dataset is constructed with law bills and not contracts. Nevertheless, it uses similar legal vocabulary, but the variety of law applications (e.g. environment, labour law) makes it of limited use for insurance applications. The fourth is MAUD, an expert-annotated merger agreement understanding dataset for reading comprehension questions about merger agreements. However, again, the dataset does not transfer well to the insurance domain. Finally, a more recent corpus is EUR-Lex-Sum, a manually curated multi- and cross-lingual document summaries of legal acts from the European Union law platform. It contains up to 1,505 document/summary pairs for 24 languages. Like BillSum, it is constructed with legal acts, thus not insurance documents.

No synthetic corpus of legal documents is available in the literature, nor are any synthetic dataset generators for legal documents. However, creating a synthetic dataset is not a new challenge. Research in many areas, such as finance, healthcare and computer vision, use synthetic datasets \cite{gursakal2023introduction}. Synthetic data generation is usually categorized into two distinct categories: process-driven methods and data-driven methods. Process-driven methods generate synthetic data from mathematical models of an underlying physical process; for example, numerical simulations using Monte Carlo. Data-driven methods generate synthetic data from generative models that have been trained on real data 
\cite{gursakal2023introduction}. Most recent approaches are data-driven and rely on generative methods using generative adversarial networks (GAN) \cite{gursakal2023introduction}. GANs are deep neural networks that produce two jointly-trained networks; one generates synthetic data intended to be as similar as possible to the training data, and one tries to discriminate the synthetic data from true training data. They have proven to be very good at learning high-dimensional, continuous data such as images \cite{gursakal2023introduction}. However, GAN data generators (or any data-driven approach) usually generate images, numerical values and short texts (i.e. sentences), not long coherent documents such as an insurance contract. 
Thus, solutions like the DataSynthetizer \cite{howe2017synthetic} or Synthetic Data Generation (SDV) \cite{7796926} Python packages that use generative methods are not well suited to generate long textual data. Neither are other solutions using large language models (LLM) \cite{bayer2022survey}. Indeed, most recent approaches using LLM as the generative method are applied on relatively short documents compared to long insurance contracts (90 to 100 pages). For example, \cite{bayer2022data} have used GPT-2 to generate new data of \guillemet{long} document of more than 280 tokens using the SST movies reviews dataset as a finetuning dataset for GPT-2. Thus, the meaning of the \guillemet{long} document is shorter than insurance contracts. Also, as \cite{brown2020language} stated, for long text, LLM tends to repeat themselves semantically at the document level, start to lose coherence over sufficiently long passages, contradict themselves and include factual inaccuracies. In other words, for now, LLMs do not show the capabilities to generate 90 to 100 pages that look like actual insurance contracts that do not include unfactual information.


\section{Realistic Insurance Synthetic Contract Data Generator}
\label{sec:risk}
As stated in \autoref{sec:intro}, insurance contracts include personal data and corporate intellectual property; for those reasons, it was impossible to publicly release a real insurance contract dataset. Therefore, in partnership with a Canadian insurance company, we propose the \guillemet{Realistic Insurance Synthetic Contract}\footnote{\href{https://github.com/GRAAL-Research/risc}{https://github.com/GRAAL-Research/risc}} (RISC) data generator, an open-source Python package to generate realistic insurance synthetic contract datasets. It was developed to be as realistic as possible by being enriched and validated by the insurer's expertise. 
RISC uses a set of templates, statistical data models, and a synthetic protection generator trained on real insurance data to create synthetic data. As a result, starting from an initial seed, it can generate a deterministic dataset of non-annotated French and English realistic synthetic automobile insurance contracts based on the AMF-approved Quebec form and the insurer documentation. Real insurance contracts are composed of the followings parts; thus, synthetic one uses the same parts:

\begin{itemize}[leftmargin=*, label={}]
    \item \textbf{Insurer introductory pages}: consists of pages that introduce the insurer (e.g. customer service phone number), table of contents, client customer advantages (e.g. privileged rates) and actions required by customers (e.g. detach and keep insurance certificate). This part is typically 4 to 5 pages long.
    \item \textbf{Declaration and disclosure}: consists of details about the insurance contract. Notably, it includes the main driver and vehicle information, contract start and end date, and contract insurance coverage. This part is typically 2 to 3 pages long.
    \item \textbf{Quebec Police Form (Q.P.F.)}: consists of the AMF-approved automobile insurance form specifying the insurer's and insured's legal obligations, including and excluding coverage of the mandatory liability coverage and the property car damage and the general conditions. The regulatory form does not cover all the regulated covered risks. Instead, it offers limited coverage. For example, the form covers the insured car but with depreciation. This part is 34 pages in French and 33 pages in English.
    \item \textbf{Quebec endorsements form (Q.E.F.)}: consists of the set of 81 possibles clauses added to the contract to increase or decrease the coverage of the base form. For example, an insurance contract can include an endorsement to cover the insured car without depreciation. In other words, endorsements \guillemet{contradict} the base form text. Endorsements are typically 1 page long, but some can go up to 10 pages. 
\end{itemize}

\begin{figure}
    \centering
    \includegraphics[scale=0.65, keepaspectratio]{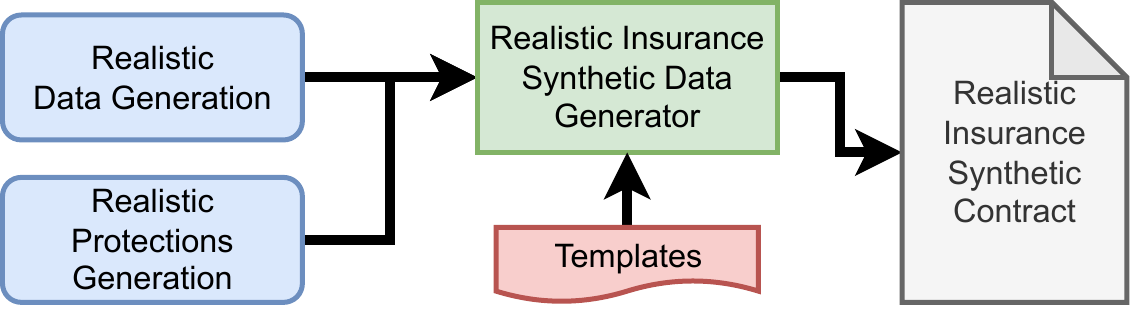}
    \captionsetup{width=\linewidth}
    \caption{Illustration of RISC procedure to generate a realistic insurance synthetic contract.}
    \vspace{-0.75em}
    \label{fig:procedure}
\end{figure}

\autoref{fig:procedure} illustrates RISC's generation procedure (\textcolor{green}{\textbf{green}}) to generate a realistic synthetic automobile insurance contract (\textcolor{gray}{\textbf{gray}}). It uses two components to generate an insurance contract: data generators (\textcolor{blue}{\textbf{blue}}) and templates (\textcolor{red}{\textbf{red}}). First, it uses template-filling templates to ensure the proper generation of the contract structure. Second, it uses two generators designed to populate the templates: a realistic protection generator and a realistic data generator. These data generators produce the synthetic information included in the insurance contract, such as names and addresses. All three components will be discussed in the following sub-sections.

\subsection{Templates}
To generate realistic insurance contracts, we have manually designed a set of fillable templates along with the generator's synthetic data based on the insurer's expertise. We created various templates in both French and English for all four parts of the insurance contract. Templates were created by manually extracting real insurance contract contents that were not insurance company information (e.g. name of the insurance company) or the insured information data (e.g. name, address, car details). Then, missing information, such as the insured name and car details, was marked as fillable data. The templates for the first two parts of the insurance contract are designed based on the insurer's corporate documentation. However, company-specific information in the documentation was depersonalized by replacing it with fake information that can be customized. For example, the \guillemet{Insurer Customer Service} phone number can be replaced by any phone number. The templates for the last two parts of the contract are designed from the approved forms available online at the AMF Website \cite{amf_form}. In total, for both languages, we created 29 templates for the first three parts of the insurance contract and 25 for the endorsements. \autoref{fig:template_example} presents an example of a template used by our synthetic generator.

\begin{figure}
    \centering
    \small
    \ttfamily
    \noindent\fbox{%
        \parbox{\textwidth}{%
            \fontseries{l}\selectfont Item 2. CONTRACT PERIOD\\
            FROM: \fontseries{b}\selectfont <Contract Start Date>\fontseries{l}\selectfont* TO: \fontseries{b}\selectfont <Contract End Date>\fontseries{l}\selectfont* EXCLUSIVELY\\
            \fontseries{l}\selectfont *at 12:01 A.M. standard time at the address of the named insured.
            \captionsetup{width=\linewidth}
        }%
    }%
    \caption{Fillable template example used by RISC to generate insurance contract.}
    \label{fig:template_example}
\end{figure}

\subsection{Realistic Protection Generation}
The objective of the realistic protection generator is to generate a set of realistic protections for an insurance contract. The protections can include liabilities (Section A) and property damage (Sections B1 to B4) coverage, and the 81 available endorsements (Q.E.F. section) that increase or decrease insurance coverage.
For each protection, a binary value represents whether or not the protection is included. \autoref{tab:example:protection} presents an example of a set of binary protections. However, these protections are not independent of each other; some build upon others, while some are mutually exclusive. Consequently, based on knowledge from our partner insurance company, we designed a set of rules to constrain how protections can interact with each other, and guarantee that the set generated corresponds to a likely insurance contract. Specifically, a set of protections must comply with the following rules to be realistic:
\begin{itemize}[leftmargin=*]
    \item Include the mandatory Section A coverage.
    \item It does not include Section B1 with any other Section B coverage, since Section B1 is a superset of all the other Section B.
    \item It does not include both Section B3 and Section B4 since Section B3 is a superset of Section B4.
    \item It does not include the Q.E.F. 41, which removes the deductible on some risk if the insured has a claim or a driver's license suspension.
    \item It does not include a Q.E.F. 43, which covers the insured car without depreciation, without any Section B coverage, since Q.E.F. 43 is a replacement value applied to property damage described in Section B.
\end{itemize}

A rules-based approach enforces these rules. That is, it generates a set of protection and verifies if these rules are respected, and if it does not, it is rejected, and the process is repeated until a set of protections respect the rules.

\begin{table}
    \centering
    \begin{tabular}{@{}ccccccccc@{}}
        \toprule
        Section A & Section B1 & Section B2 & Section B3 & Section B4 & Q.E.F. 2 & Q.E.F. 3 & $\hdots$ & Q.E.F. 48a \\ \midrule
        1 & 0 & 1 & 1 & 0 & 0 & 1 & $\hdots$ & 0 \\ \bottomrule
    \end{tabular}
    \captionsetup{width=\linewidth}
    \caption{Example of a set of protections for a Quebec insurance contract, provided by the insurer.}
    \label{tab:example:protection}
    \vspace{-0.5em}
\end{table}

The insurance company provided us with a real insurance tabular dataset to develop a synthetic protection generator that can generate realistic data. This dataset consists of 266,082 binary protections similar to the one shown in \autoref{tab:example:protection}. However, since insurers are not required to cover all 81 endorsements, our dataset includes only the 26 endorsements covered by our partner. Based on the insurer dataset, on average, an insurance contract (a row) includes 7.24 protections, including mandatory civil liability, and all the contracts include at least one endorsement. Moreover, as shown in \autoref{tab:uc}, there are 1,880 unique combinations of protection (a set of columns), and 75~\% of them appear at most in 0.00004~\% of the dataset. This means that using the unique combination's distribution to generate a synthetic protection dataset would be cumbersome due to many rarely-occurring combinations. Furthermore, such an approach would only generate a combination of protections seen during training. The insurer was also unwilling to share a model to generate a perfect distribution of its risk portfolio. Thus, a look-alike distribution was more suitable for a public dataset.

\begin{table}
    \centering
    \begin{tabular}{@{}ccccc@{}}
        \toprule
        \begin{tabular}[c]{@{}c@{}}Unique combination \\ (UC)\end{tabular} & \begin{tabular}[c]{@{}c@{}}Average UC\\ frequency (\%)\end{tabular} & \begin{tabular}[c]{@{}c@{}}UC frequency\\ median (\%)\end{tabular} & \begin{tabular}[c]{@{}c@{}}UC frequency \\ 75-quartile (\%)\end{tabular} & \begin{tabular}[c]{@{}c@{}}Maximum UC\\frequency (\%)\end{tabular} \\ \midrule
        1,880 & 0.00053 & 0.00001 & 0.00004 & 0.12872 \\ \bottomrule
    \end{tabular}
    \captionsetup{width=\linewidth}
    \caption{Distribution of the unique combinations of the insurer protection dataset.}
    \label{tab:uc}
    \vspace{-1em}
\end{table}

Since the data to generate are composed of numerical values, we have trained a tabular variational autoencoder (TVAE) and a conditional tabular GAN (CTGAN) model using \cite{ctgan}'s approach. The TVAE model uses a modified version of the traditional VAE loss function to adapt to tabular data. The CTGAN model is a conditional GAN for synthetic tabular data generation using mode-specific normalization. The advantage of using these approaches is that they rely on a neural network generative model to capture the relationship between the distributions of a specific protection (a column) and all the other protections. For example, it is common to see a \guillemet{bundle} of endorsements purchased together, such as Q.E.F. 20a and 27, to cover civil liability for a short rental car during a vacation trip. These approaches capture commonly-occurring sets of protections but do not restrict the generative model to generate data seen during training. Therefore, the data will be realistic but will differ slightly from the insurer's portfolio risk.

To train our two models, we use the SDV \cite{7796926} implementation of TVAE and CTGAN models. We train each of the aforementioned models using the random initial seed 42 with a batch size of 1,024. The models were trained for 200 epochs using SDV default training parameters for the generator and discriminator dimensions and learning rate. The training was done using the entire dataset since SDV evaluates models by comparing the quality of a synthetic sampled test dataset to the original one. It does so by computing the inverted Kolmogorov-Smirnov (KS) test \cite{massey1951kolmogorov} between the two datasets. We have used a synthetic sampled test dataset size of 300,000. \autoref{tab:res} shows the averaged metric values for both models. These results show that both models achieved high scores on the KS test, but the TVAE model slightly outperformed the CTGAN model. We conducted a z-test significance test on both models' KS test scores to further assess the models' performance. Our z-test null hypothesis is that the pair of models have equal performances, meaning that values smaller or greater than $|3.290527|$ allow us to reject the null hypothesis with $\alpha = 0.001$. A positive value means that the first model (left) performs significantly better than the second (right), and a negative value means the opposite. The z-test value is 70.63, so we can reject the null hypothesis that both models share the same performance. It also means that the TVAE performs significantly better than the CTGAN model. Second, both models create synthetic data with similar unique combination distributions as the insurer dataset. Third, the CTGAN tends to generate nearly double the number of new unique combinations (UC) of a set of protections, with 1,842 of them being entirely new (not seen during training). Conversely, TVAE creates more look-alike protections by generating more sets of protections similar to the real data. Therefore, since TVAE has significantly better performance, is less computationally intensive, easier to use and tends to offer more look-alike protections to the insurer dataset, we selected this model as the protection generation model.

\begin{table}
    \centering
    \resizebox{\textwidth}{!}{%
    \begin{tabular}{@{}lccccccc@{}}
        \toprule
        &\begin{tabular}[c]{@{}c@{}}Inverted\\ KS test\end{tabular}&\begin{tabular}[c]{@{}c@{}}Unique \\ combination \\ (UC)\end{tabular} &\begin{tabular}[c]{@{}c@{}}New\\UC\end{tabular} & \begin{tabular}[c]{@{}c@{}}Average UC\\ frequency\\ (\%)\end{tabular} & \begin{tabular}[c]{@{}c@{}}UC frequency\\ median\\ (\%)\end{tabular} & \begin{tabular}[c]{@{}c@{}}UC frequency \\ 75-quartile\\ (\%)\end{tabular} & \begin{tabular}[c]{@{}c@{}}Maximum UC \\ frequency\\ (\%)\end{tabular} \\ \midrule
        Insurer data & - & 1,880 & - &0.00053 & 0.00001 & 0.00004 & 0.12872\\
        TVAE & 0.9964 & 1,605 & 535 & 0.00062 & 0.00001 & 0.00007 & 0.12689\\
        CTGAN & 0.9746 & 2,912 & 1,842 & 0.00034 & 0.00001 & 0.00003 & 0.11602\\
        \bottomrule
    \end{tabular}%
    }
    \captionsetup{width=\linewidth}
    \caption{Distribution analysis of the unique combination of the synthetic protection generator.}
    \label{tab:res}
    \vspace{-1.25em}
\end{table}

\subsection{Realistic Data Generation}
The objective of the realistic data generator is to generate a set of data similar to those in a real insurance contract. However, since most of these data include personal information such as date of birth, address, car details, and driving record, it is impossible to use real data to develop a synthetic data generator due to confidentiality concerns, unlike the realistic protection generator. Hence, using our and the insurer's expertise, we have selected a mix of preset statistic generators available in the literature and crafted stochastic generators to compose the realistic data generator; they are listed below:

\begin{itemize}[leftmargin=*, label={}]
    \item \textbf{Insured personal information}:
    For most of the insured person's data, such as the name, address, date of birth, unique client ID, and association rebate, we have used the Python Faker library \cite{Faraglia_Faker}. It uses preset data to sample fake data randomly. For example, to generate names, Faker uses a preset of first and last names and samples in both presets to create a completely fake name. For the sex, we have used stochastic sampling using realistic distribution parameters based on the driver population presented in the 2021 SAAQ road safety record \cite{saaq_2021_report}.
    \item \textbf{
    Insured driving information}: For the insured person's driving information, namely the number of claims in the past five years and the number of driving suspensions, we have used stochastic sampling using realistic distribution parameters based on the past eleven years' GAA Quebec's claims data \cite{gaa_claims} and the 2019 SAAQ driver suspension data \cite{saaq_2019_report}. We have chosen the 2019 SAAQ driver suspension data to avoid the COVID restrictions of 2020-2021, when license suspensions significantly dropped due to reduced opportunities to drive (and thus to be caught in a driving infraction by police and receive a suspension).
    \item \textbf{Protections coverage amount}: For the protection coverage amounts of the liability coverage and the property damage deductible, we have used stochastic sampling using realistic distribution parameters based on the insurer's expertise.
    \item \textbf{Vehicle information}: To generate the vehicle data (e.g. year, maker, model, motor type (e.g. electric) and financing institution details), we use the Python Faker library. For the purchase condition, we use a stochastic sampling using realistic distribution parameters based on the 2022 Statistics Canada quarterly new motor vehicle registrations \cite{stats_can} and 2021 SAAQ road safety record \cite{saaq_2021_report}.
    \item \textbf{Contract information}: The contract starting date is generated using the Python Faker library in the range of up to one year before the generation date. For the contract premium details per protection, we use stochastic sampling from realistic distribution parameters based on the insurer's expertise and the 2021 GAA premium statistics \cite{gaa}.
\end{itemize}

In order to reduce the complexity of the data generation process, we also designed the system to only generate data for one-year contracts of new customers that cover a single insured person on a single car. These represent the most common type of car insurance contract. However, this limitation can easily be removed if a more general insurance dataset needs to be generated.

\section{Realistic Insurance Synthetic Bilingual Automobile Contract Dataset}
\label{sec:rsbaic}
We created the Realistic Insurance Synthetic Bilingual Automobile Contract (RISCBAC) dataset\footnote{\href{https://huggingface.co/datasets/davebulaval/RISCBAC}{https://huggingface.co/datasets/davebulaval/RISCBAC}} using RISC to enable ML research in the insurance field. It consists of 10,000 French and English realistic synthetic automobile insurance contracts. The dataset is generated using the initial seed 42 for each language. As a result, the contracts in both datasets have the same protections and data.

\subsection{Datasets Analysis}
\autoref{tab:datasets_statistics} presents some key statistics of French and English RISCBAC lower-cased datasets, and the legal corpora introduce in \autoref{sec:rel_work}. For the legal corpora, we have used their official version on the \guillemet{HuggingFace Datasets Hub}\footnote{\href{https://huggingface.co/datasets}{https://huggingface.co/datasets}}, except for LEDGAR, which was not available. Instead, we have used LEDGAR's official clean version available online\footnote{\href{https://drive.switch.ch/index.php/s/j9S0GRMAbGZKa1A}{https://drive.switch.ch/index.php/s/j9S0GRMAbGZKa1A}}. 
For each of these corpora, depending on the dataset type (i.e. the task), we kept only the \guillemet{(\textit{colunm name})} written below the dataset name shown in \autoref{tab:datasets_statistics}. 
For example, for the BillSum dataset, we only kept the \guillemet{\textit{text}} column, thus excluded the \guillemet{\textit{summary}} and \guillemet{\textit{title}} from the statistics.
All statistics were computed using SpaCy \cite{Honnibal_spaCy_Industrial-strength_Natural_2020}, and they excluded new line (\verb|\n|), whitespace, punctuation and some special characters (\verb|<|, \verb|>|, \texttt{|} and \verb|$|), and numeric character tokens. We will first analyze English and French RISCBAC datasets in the following two sub-sections and then compare them with other legal corpora using \autoref{tab:datasets_statistics}.

\subsubsection{RISCBAC Datasets Comparison}
First, we can see in \autoref{tab:datasets_statistics} that the datasets in both languages share a relatively similar number of tokens and lexical words (LW) (i.e. non-stopwords), with French having only 11\% more tokens than English. Second, the vocabulary size is relatively small since all insurance contracts share the same base contract and only vary in endorsements and data (e.g. insured name and address). However, we note that English has 66~\% more vocabulary than French. Third, documents are long; they include, on average, 1,071 and 996 sentences in 98 and 95 pages. Fourth, we can see that the documents are complex. They, on average, are composed of wordy sentences (25 tokens long). For example, the UK government's best writing practices policy stated that official publications should not use sentences of more than 25 words and use an average of 14 words \cite{govuk}. Finally, to evaluate the reading complexity level of the contracts, we compute readability scores using the following three frequently used formulas: Flesch-Kincaid \cite{flesch1948readability}, Gunning fog index \cite{gunning1969fog} and SMOG \cite{mc1969smog}. They compute using a scale from 0 (hardest) to 100 (easier) to assess the readability level. All formulas use slightly different approaches to measure the difficulty level. We can see that the two contracts datasets score near minimal on all three metrics, making them very complicated to read.

\subsubsection{RISCBAC Comparison With Other Legal Corpora}
Refering again to \autoref{tab:datasets_statistics}, RISCBAC datasets contains much longer documents than any other dataset, with nearly double the number of tokens and 150~\% more sentences per document compared to the second-longest-documents in the EUR-Lex-Sum. 
On the other hand, RISCBAC sentences are among the shortest in the table, nearly five times shorter than the maximum found in MAUD, and have the lowest lexical richness.  
Despite this, RISCBAC documents achieve the lowest Flesch-Kincaid readability score, demonstrating that insurance contracts are longer and more complicated to read than other legal documents. These results highlight how 
insurance contracts are a very different and much more complex type of document than those found in traditional NLP corpora and even legal NLP corpora.


\begin{table}
    \resizebox{\textwidth}{!}{%
    \begin{tabular}{@{}l|cc|cccccc@{}}
        \toprule
         & \begin{tabular}[c]{@{}c@{}}RISCBAC\\ French\end{tabular} & \begin{tabular}[c]{@{}c@{}}RISCBAC\\ English\end{tabular} & \begin{tabular}[c]{@{}c@{}}LEDGAR\\ (provision)\end{tabular} & \begin{tabular}[c]{@{}c@{}}CUAD\\ (context)\end{tabular} & \begin{tabular}[c]{@{}c@{}}BillSum\\ (text)\end{tabular} & \begin{tabular}[c]{@{}c@{}}MAUD\\ (text)\end{tabular} & \begin{tabular}[c]{@{}c@{}}EUR-Lex-Sum\\ French\\ (reference)\end{tabular} & \begin{tabular}[c]{@{}c@{}}EUR-Lex-Sum\\ English\\ (reference)\end{tabular} \\ \midrule
        Number of documents & 10,000 & 10,000 & 846,274 & 26,632 & 23,455 & 39,231 & 1,505 & 1,504 \\
        Vocabulary size & 19,159 & 31,869 & 79,582 & 38,722 & 120,683 & 6,130 & 226,558 & 218,835 \\
        Avg number of tokens & 26,869.85 & 24,198.49 & 122.45 & 9,092.28 & 1,271.22 & 450.99 & 14,484.40 & 12,636.66 \\
        Avg number of LW & 13,109.94 & 12,968.63 & 59.24 & 4,932.46 & 707.94 & 231.19 & 7,388.66 & 7,132.57 \\
        Avg number of sentence & 1,070.88 & 996.35 & 2.11 & 264.52 & 52.36 & 4.04 & 714.47 & 399.68 \\
        \begin{tabular}[c]{@{}l@{}}Avg sentence length\\ (tokens)\end{tabular} & 25.09 & 24.40 & 63.67 & 36.43 & 26.46 & 163.89 & 60.40 & 45.38 \\
    \begin{tabular}[c]{@{}l@{}}Avg sentence length\\ (LW)\end{tabular} & 12.34 & 13.13 & 30.71 & 19.82 & 14.72 & 83.69 & 30.19 & 25.15 \\
    Avg number of pages & 98.05 & 95.05 & N/A & N/A & N/A & N/A & N/A & N/A \\
    Lexical richness & 0.00014 & 0.00024 & 0.00158 & 0.00029 & 0.00725 & 0.00065 & 0.02034 & 0.02037 \\
    Avg Flesch-Kincaid score & 11.73 & 13.77 & 25.60 & 16.40 & 15.76 & 61.77 & 19.45 & 19.58 \\ 
    Avg Gunning fog score & 10.81 & 10.47 & 27.65 & 15.04 & 14.98 & 63.09 & 18.74 & 17.42\\
    Avg SMOG score & 14.18 & 15.97 & 6.82 & 16.65 & 16.73 & 15.32 & 17.86 & 19.42\\
    \bottomrule
    \end{tabular}%
    }
    \captionsetup{width=\linewidth}
\caption{Aggregate statistics of the RISCBAC datasets and legal corpora introduce in \autoref{sec:rel_work}.}
\label{tab:datasets_statistics}
\end{table}

\subsection{Research using RISCBAC}
In this section, we discuss ML NLP tasks that can be performed on the RISCBAC dataset and those tasks that require additional work on the dataset before it can be used.

The documents generated can be used for research on unsupervised automatic text summarization
\cite{al2021automated}, unsupervised question answering 
\cite{martinez2021survey} and unsupervised information retrieval 
\cite{sansone2022legal}, unsupervised legal text simplification 
\cite{garimella2022text}, unsupervised machine translation 
\cite{gibadullin2019survey}, text anonymization 
\cite{csanyi2021challenges}, and coreference resolution of clauses \cite{stolfo-etal-2022-simple, 10.1007/978-3-030-96957-8_25}. In addition, it could also be used as a low-resource dataset for meta-learning tasks
\cite{yin2020meta}. The unique features of insurance contracts make our RISCBAC dataset particularly interesting for these tasks compared to other available datasets. Working with such lengthy documents is challenging due to the computing limitations of current state-of-the-art deep learning methods such as Transformer 
\cite{10.1145/3545176}. Furthermore, as stated in the \autoref{sec:intro}, insurance contracts \guillemet{contradict} themselves between the base form and the endorsements. As a result, tasks such as summarization, information retrieval and question-answering become more challenging. Few works focus on handling contradictions in sentences \cite{krishna2022similarity}, and even fewer in documents, with most of them focusing on misinformation detection \cite{wu2022cross}, or multi-document contradictions \cite{ma2022multi}. The contradictions found in our dataset are of a different and much more challenging nature. 

Furthermore, the RISCBAC dataset can also be used for research on tasks such as legal named entity recognition (NER) 
\cite{10.1007/978-3-030-33220-4_20}, supervised machine translation \cite{gibadullin2019survey}, supervised coreference document resolution \cite{https://doi.org/10.48550/arxiv.2006.05621} and contract element extraction 
\cite{10.1145/3086512.3086515}. However, doing so will require further annotations of the dataset. Annotations must be provided and validated for each specific task to use the corpus to train supervised ML algorithms. 
For instance, for the NER task, it would require annotating relevant named entities such as the insured name, address, car details, and named law article and contract Item (e.g. Item 3, Civil Code Art. 2). For supervised machine translation, it would require to do a pre-processing text alignment 
\cite{bott-saggion-2011-unsupervised}. The supervised coreference document resolution would require manual or semi-manual annotation of a specific portion of a document referring to another portion of the insurance contract. Finally, the contract element extraction would require manual annotation of relevant element extraction similar to the NER data but also including contract elements such as items and clauses.

\section{Conclusion}
This paper presented RISC, an open-source Python package we created to generate realistic synthetic insurance contracts. It is designed to mimic Quebec's automobile insurance contracts.
We also presented RISCBAC, a realistic bilingual synthetic automobile insurance contract dataset. The dataset currently comprises 10,000 French and English synthetic automobile insurance contracts in \verb|.txt| format. Both contributions are designed to enable NLP experiments applied to insurance documents, a very different and much more difficult class of documents than those in traditional NLP corpora. 

To continue our work, we aim to extend the type of insurance documents RISC can generate to include residential property and collective insurance. Unlike automotive insurance contracts, these contracts do not have a mandatory regulated form in Quebec and Canada, but rather a variable \guillemet{standard form} and, moreover, are primarily proprietary documents. We also aim to include an automatic annotation step of named entities during the RISC generation process.

\section*{Acknowledgements}
This research was made possible thanks to the support of a Canadian insurance company, NSERC research grant RDCPJ 537198-18 and FRQNT doctoral research grant. We wish to thank the reviewers for their comments regarding our work.

\printbibliography[heading=subbibintoc]

@inproceedings{ctgan,
  title={{Modeling Tabular Data Using Conditional GAN}},
  author={Xu, Lei and Skoularidou, Maria and Cuesta-Infante, Alfredo and Veeramachaneni, Kalyan},
  booktitle={Advances in Neural Information Processing Systems},
  year={2019}
}

@inproceedings{
    7796926,
    author={N. {Patki} and R. {Wedge} and K. {Veeramachaneni}},
    booktitle={International Conference on Data Science and Advanced Analytics},
    title={{The Synthetic Data Vault}},
    year={2016},
    pages={399-410},
    ISSN={},
    month={10}
}

@article{massey1951kolmogorov,
  title={{The Kolmogorov-Smirnov Test for Goodness of Fit}},
  author={Massey Jr, Frank J},
  journal={Journal of the American statistical Association},
  volume={46},
  number={253},
  pages={68--78},
  year={1951},
  publisher={Taylor \& Francis}
}

@software{Faraglia_Faker,
    author = {Faraglia, Daniele and {Other Contributors}},
    license = {MIT},
    title = {{Faker: A Python Package That Generates Fake Data for You}},
    url = {https://github.com/joke2k/faker}
}

@misc{gaa, 
    title={{Statistics: At a Glance}}, 
    author={{Groupement des assureurs automobiles}},
    url={https://gaa.qc.ca/en/statistics/at-a-glance},
    urldate = {2023-01-31},
}

@misc{gaa_claims, 
    title={{Statistics: Claims experience}}, 
    author={{Groupement des assureurs automobiles}},
    url={https://gaa.qc.ca/en/statistics/claims-experience/collision-and-upset},
    urldate = {2023-01-31},
}

@report{saaq_2021_report, 
    title={{Bilan routier, parc automobile et permis de conduire}}, 
    url={https://saaq.gouv.qc.ca/fileadmin/documents/publications/espace-recherche/dossier-statistique-2021-bilan-routier.pdf},
    author={{Société de l'assurance automobile du Québec}},
    year={2021}
}

@report{saaq_2019_report, 
    title={{Données et statistiques}}, 
    url={https://saaq.gouv.qc.ca/fileadmin/documents/publications/donnees-statistiques-2019.pdf},
    author={{Société de l'assurance automobile du Québec}},
    year={2019}
}

@misc{stats_can, 
    title={{New Motor Vehicle Registrations: Quarterly Data}}, 
    author={{Statistics Canada}},
    url={https://www150.statcan.gc.ca/n1/pub/71-607-x/71-607-x2021019-fra.htm},
    urldate = {2023-01-31}
}

@misc{amf_form, 
    title={{AMF approved forms}}, 
    author={{Autorité des marchés financiers}},
    url={https://lautorite.qc.ca/en/professionals/insurers/automobile-insurance/amf-approved-forms},
    urldate = {2023-01-31}
}

@article{csanyi2021challenges,
  title={{Challenges and Open Problems of Legal Document Anonymization}},
  author={Cs{\'a}nyi, Gergely M{\'a}rk and Nagy, D{\'a}niel and V{\'a}gi, Ren{\'a}t{\'o} and Vad{\'a}sz, J{\'a}nos P{\'a}l and Orosz, Tam{\'a}s},
  journal={Symmetry},
  volume={13},
  number={8},
  pages={1490},
  year={2021},
  publisher={MDPI}
}

@inproceedings{bott-saggion-2011-unsupervised,
    title = "An Unsupervised Alignment Algorithm for Text Simplification Corpus Construction",
    author = "Bott, Stefan  and
      Saggion, Horacio",
    booktitle = "Proceedings of the Workshop on Monolingual Text-To-Text Generation",
    month = jun,
    year = "2011",
    address = "Portland, Oregon",
    publisher = "ACL",
    url = "https://aclanthology.org/W11-1603",
    pages = "20--26",
}

@inproceedings{10.1145/3086512.3086515,
    author = {Chalkidis, Ilias and Androutsopoulos, Ion and Michos, Achilleas},
    title = {{Extracting Contract Elements}},
    year = {2017},
    isbn = {9781450348911},
    publisher = {Association for Computing Machinery},
    url = {https://doi.org/10.1145/3086512.3086515},
    doi = {10.1145/3086512.3086515},
    booktitle = {Proceedings of the ICAIL},
    pages = {19–28},
    numpages = {10},
}

@InProceedings{10.1007/978-3-030-96957-8_25,
    author="Zhukova, Anastasia
    and Hamborg, Felix
    and Donnay, Karsten
    and Gipp, Bela",
    editor="Smits, Malte",
    title="XCoref: Cross-document Coreference Resolution in the Wild",
    booktitle="Inf. for a Better World: Shaping the Global Future",
    year="2022",
    publisher="Springer Int. Publishing",
    address="Cham",
    pages="272--291",
    isbn="978-3-030-96957-8"
}

@inproceedings{stolfo-etal-2022-simple,
    title = "A Simple Unsupervised Approach for Coreference Resolution using Rule-based Weak Supervision",
    author = "Stolfo, Alessandro  and
      Tanner, Chris  and
      Gupta, Vikram  and
      Sachan, Mrinmaya",
    booktitle = "Proceedings of the Joint Conference on Lexical and Computational Semantics",
    month = jul,
    year = "2022",
    address = "Seattle, Washington",
    publisher = "ACL",
    url = "https://aclanthology.org/2022.starsem-1.7",
    pages = "79--88",
}

@misc{https://doi.org/10.48550/arxiv.2006.05621,
  doi = {10.48550/ARXIV.2006.05621},
  url = {https://arxiv.org/abs/2006.05621},
  author = {Lebanoff, Logan and Muchovej, John and Dernoncourt, Franck and Kim, Doo Soon and Wang, Lidan and Chang, Walter and Liu, Fei},
  title = {{Understanding Points of Correspondence between Sentences for Abstractive Summarization}},
  
  publisher = {arXiv:2006.05621},
  
  year = {2020},
  
  copyright = {arXiv.org perpetual, non-exclusive license}
}

@InProceedings{10.1007/978-3-030-33220-4_20,
author="Leitner, Elena
and Rehm, Georg
and Moreno-Schneider, Julian",
editor="Acosta, Maribel
and Cudr{\'e}-Mauroux, Philippe
and Maleshkova, Maria
and Pellegrini, Tassilo
and Sack, Harald
and Sure-Vetter, York",
title="Fine-Grained Named Entity Recognition in Legal Documents",
booktitle="Semantic Systems. The Power of AI and Knowledge Graphs",
year="2019",
publisher="Springer International Publishing",
address="Cham",
pages="272--287",
isbn="978-3-030-33220-4"
}

@inproceedings{Eidelman_2019, 
	url = {https://aclanthology.org/D19-5406},
  
	year = 2019,
	publisher = {ACL},
  
	author = {Vladimir Eidelman},
  
	title = {{BillSum: A Corpus for Automatic Summarization of US Legislation}},
  
	booktitle = {Proceedings of the Workshop on New Frontiers in Summarization}
}

@misc{https://doi.org/10.48550/arxiv.2103.06268, 
  url = {https://arxiv.org/abs/2103.06268},
  
  author = {Hendrycks, Dan and Burns, Collin and Chen, Anya and Ball, Spencer},
  
  title = {{CUAD: An Expert-Annotated NLP Dataset for Legal Contract Review}},
  
  publisher = {arXiv:2103.06268},
  
  year = {2021},
}

@article{katz2023natural,
  title={{Natural Language Processing in the Legal Domain}},
  author={Katz, Daniel Martin and Hartung, Dirk and Gerlach, Lauritz and Jana, Abhik and Bommarito, Michael James},
  journal={Available at SSRN 4336224},
  year={2023}
}

@article{chalkidis2021lexglue,
  title={{LexGLUE: A Benchmark Dataset for Legal Language Understanding in English}},
  author={Chalkidis, Ilias and Jana, Abhik and Hartung, Dirk and Bommarito, Michael and Androutsopoulos, Ion and Katz, Daniel Martin and Aletras, Nikolaos},
  journal={arXiv:2110.00976},
  year={2021}
}

@article{wang2023maud,
  title={{MAUD: An Expert-Annotated Legal NLP Dataset for Merger Agreement Understanding}},
  author={Wang, Steven H and Scardigli, Antoine and Tang, Leonard and Chen, Wei and Levkin, Dimitry and Chen, Anya and Ball, Spencer and Woodside, Thomas and Zhang, Oliver and Hendrycks, Dan},
  journal={arXiv:2301.00876},
  year={2023}
}

@article{beauchemin2020generating,
  title={{Generating Intelligible Plumitifs Descriptions: Use Case Application With Ethical Considerations}},
  author={Beauchemin, David and Garneau, Nicolas and Gaumond, Eve and D{\'e}ziel, Pierre-Luc and Khoury, Richard and Lamontagne, Luc},
  journal={arXiv:2011.12183},
  year={2020}
}

@article{parada2020digital,
  title={{Digital Court Records: a Diversity of Uses}},
  author={Parada, Alexandra and Prom Tep, Sandrine and Millerand, Florence and Noreau, Pierre and Santorineos, Anne-Marie},
  journal={The Annual Review of Interdisciplinary Justice Research Volume 9},
  pages={141},
  year={2020}
}

@article{tep2019legal,
  title={{Legal Information in Digital Form: the Challenge of Accessing Computerized Court Records}},
  author={Tep, Sandrine Prom and Millerand, Florence and Parada, Alexandra and Bahary, Alexandra and Noreau, Pierre and Santorineos, Anne-Marie},
  journal={The Annual Review of Interdisciplinary Justice Research Volume 8},
  pages={217},
  year={2019}
}

@article{susskind2019online,
  title={{Online Courts and the Future of Justice}},
  author={Susskind, Richard},
    journal = {Oxford University Press},
    year = {2019},
    month = {11},
    url = {https://doi.org/10.1093/oso/9780198838364.001.0001},
}

@book{barton2017rebooting,
  title={{Rebooting Justice: More Technology, Fewer Lawyers, and the Future of Law}},
  author={Barton, Benjamin H and Bibas, Stephanos},
  year={2017},
  publisher={Encounter Books}
}

@inproceedings{rusakova2021challenges,
  title={{Challenges of the Judicial Systems of the Russian Federation and People’s Republic of China in the Era of the Pandemic}},
  author={Rusakova, Ekaterina P and Frolova, Evgenia E and Arzumanova, Lana L},
  booktitle={Modern Global Economic System: Evolutional Development vs. Revolutionary Leap 11},
  pages={1541--1549},
  year={2021},
  organization={Springer}
}

@misc{matyas2021imagining,
  title={{Imagining Resilient Courts: From COVID to the Future of Canada's Judicial System}},
  author={Matyas, David and Wills, Peter and Dewitt, Barry},
  url={ https://ssrn.com/abstract=3778869},
  year={2021}
}

@article{choi2023chatgpt,
  title={{ChatGPT Goes to Law School}},
  author={Choi, Jonathan H and Hickman, Kristin E and Monahan, Amy and Schwarcz, Daniel},
  journal={Available at SSRN},
  year={2023}
}

@article{bommarito2022gpt,
  title={{GPT Takes the Bar Exam}},
  author={Bommarito II, Michael and Katz, Daniel Martin},
  journal={arXiv:2212.14402},
  year={2022}
}

@misc{bill141,
        author                    =       {{National Assembly of Québec}},
        publisher                       =       {{National Assembly of Québec}},
        title                           =       {{An Act mainly to improve the regulation of the financial sector, the protection of deposits of money and the operation of financial institutions}},
        year                            =       2018,
        }

@misc{loiamf,
        author                    =       {{Recueil des lois et des règlements du Québec (RLRQ)}},
        publisher                       =       {Recueil des lois et des règlements du Québec},
        title                           =       {{Act Respecting the Regulation of the Financial Sector}},
        year                            =       2004,
        }

@misc{RCCAQ,
	title = {{Projet de loi 141 et vente par internet: où en est le RCCAQ?}},
	note = {\url{https://www.rccaq.com/cgi/page.cgi/_article_fr.html/Categories/Dans_la_mire/Projet_de_loi_141_et_vente_par_internet_o_en_est_le_RCCAQ_}},
	year={2018},
	month = {1},
	author={Johnson, Christopher}
}

@book{memoireamfloi141,
	author = {{Autorité des marchés financiers (AMF)}},
	year = {2018},
	title = {Mémoire présenté à la Commission des finances publiques sur le Projet de loi 141 : Loi visant principalement à améliorer l’encadrement du secteur financier, la protection des dépôts d’argent et le régime de fonctionnement des institutions financières},
	url = {https://lautorite.qc.ca/fileadmin/lautorite/grand_public/publications/professionnels/assemblee-nationale/20180118-memoire-pl141.pdf},
	publisher = {Autorité des marchés financiers}
}

@inproceedings{tuggener2020ledgar,
  title={{LEDGAR: A Large-Scale Multi-Label Corpus for Text Classification of Legal Provisions in Contracts}},
  author={Tuggener, Don and Von D{\"a}niken, Pius and Peetz, Thomas and Cieliebak, Mark},
  booktitle={Proceedings of the Language Resources and Evaluation Conference},
  pages={1235--1241},
  year={2020}
}

@article{10.1145/3545176,
author = {Koh, Huan Yee and Ju, Jiaxin and Liu, Ming and Pan, Shirui},
title = {{An Empirical Survey on Long Document Summarization: Datasets, Models, and Metrics}},
year = {2022},
issue_date = {August 2023},
publisher = {Association for Computing Machinery},
address = {New York, NY, USA},
volume = {55},
number = {8},
issn = {0360-0300},
url = {https://doi.org/10.1145/3545176},
month = {12},
articleno = {154},
numpages = {35}
}

@inproceedings{wu2022cross,
  title={{Cross-Document Misinformation Detection Based on Event Graph Reasoning}},
  author={Wu, Xueqing and Huang, Kung-Hsiang and Fung, Yi and Ji, Heng},
  booktitle={Proceedings of the Conference of the North American Chapter of the ACL: Human Language Technologies},
  pages={543--558},
  year={2022}
}

@article{krishna2022similarity,
  title={{Similarity of Sentences With Contradiction Using Semantic Similarity Measures}},
  author={Krishna Siva Prasad, M and Sharma, Poonam},
  journal={The Computer Journal},
  volume={65},
  number={3},
  pages={701--717},
  year={2022},
  publisher={Oxford University Press}
}

@article{ma2022multi,
  title={{Multi-Document Summarization via Deep Learning Techniques: A Survey}},
  author={Ma, Congbo and Zhang, Wei Emma and Guo, Mingyu and Wang, Hu and Sheng, Quan Z},
  journal={ACM Comp. Surveys},
  volume={55},
  number={5},
  pages={1--37},
  year={2022},
  publisher={ACM New York, NY}
}

@incollection{gursakal2023introduction,
  title={{An Introduction to Synthetic Data}},
  author={G{\"u}rsakal, Necmi and {\c{C}}elik, Sadullah and Biri{\c{s}}{\c{c}}i, Esma},
  booktitle={Synthetic Data for Deep Learning: Generate Synthetic Data for Decision Making and Applications with Python and R},
  pages={1--29},
  year={2023},
  publisher={Springer}
}

@article{howe2017synthetic,
  title={{Synthetic Data for Social Good}},
  author={Howe, Bill and Stoyanovich, Julia and Ping, Haoyue and Herman, Bernease and Gee, Matt},
  journal={arXiv:1710.08874},
  year={2017}
}

@article{Honnibal_spaCy_Industrial-strength_Natural_2020,
author = {Honnibal, Matthew and Montani, Ines and Van Landeghem, Sofie and Boyd, Adriane},
title = {{SpaCy: Industrial-strength Natural Language Processing in Python}},
year = {2020}
}

@article{flesch1948readability,
  title={{A Readability Formula in Practice}},
  author={Flesch, Rudolf},
  journal={Elementary English},
  volume={25},
  number={6},
  year={1948}
}

@misc{govuk, 
    title={{Sentence Length: Why 25 Words Is Our Limit}}, 
    author={{Government of the United Kingdom}},
    url={https://insidegovuk.blog.gov.uk/2014/08/04/sentence-length-why-25-words-is-our-limit},
    urldate = {2023-01-31},
}

@article{bayer2022survey,
  title={{A Survey on Data Augmentation for Text Classification}},
  author={Bayer, Markus and Kaufhold, Marc-Andr{\'e} and Reuter, Christian},
  journal={ACM Computing Surveys},
  volume={55},
  number={7},
  pages={1--39},
  year={2022},
  publisher={ACM New York, NY}
}

@article{bayer2022data,
  title={{Data Augmentation in Natural Language Processing: A Novel Text Generation Approach for Long and Short Text Classifiers}},
  author={Bayer, Markus and Kaufhold, Marc-Andr{\'e} and Buchhold, Bj{\"o}rn and Keller, Marcel and Dallmeyer, J{\"o}rg and Reuter, Christian},
  journal={{Int. Jour. of ML and Cybernetics}},
  pages={1--16},
  year={2022},
  publisher={Springer}
}

@article{brown2020language,
  title={{Language Models Are Few-Shot Learners}},
  author={Brown, Tom and Mann, Benjamin and Ryder, Nick and Subbiah, Melanie and Kaplan, Jared D and Dhariwal, Prafulla and Neelakantan, Arvind and Shyam, Pranav and Sastry, Girish and Askell, Amanda and others},
  journal={Advances in neural information processing systems},
  volume={33},
  pages={1877--1901},
  year={2020}
}

@article{al2021automated,
  title={{Automated Text Simplification: A Survey}},
  author={Al-Thanyyan, Suha S and Azmi, Aqil M},
  journal={ACM Computing Surveys},
  volume={54},
  number={2},
  pages={1--36},
  year={2021},
  publisher={ACM New York, NY, USA}
}

@article{martinez2021survey,
  title={{A Survey on Legal Question Answering Systems}},
  author={Martinez-Gil, Jorge},
  journal={arXiv:2110.07333},
  year={2021}
}

@inproceedings{garimella2022text,
  title={{Text Simplification for Legal Domain:$\{$I$\}$ nsights and Challenges}},
  author={Garimella, Aparna and Sancheti, Abhilasha and Aggarwal, Vinay and Ganesh, Ananya and Chhaya, Niyati and Kambhatla, Nanda},
  booktitle={Proceedings of the Natural Legal Language Processing Workshop},
  pages={296--304},
  year={2022}
}

@article{sansone2022legal,
  title={{Legal Information Retrieval Systems: State-Of-The-Art and Open Issues}},
  author={Sansone, Carlo and Sperl{\'\i}, Giancarlo},
  journal={Information Systems},
  volume={106},
  pages={101967},
  year={2022},
  publisher={Elsevier}
}

@article{gibadullin2019survey,
  title={{A Survey of Methods to Leverage Monolingual Data in Low-Resource Neural Machine Translation}},
  author={Gibadullin, Ilshat and Valeev, Aidar and Khusainova, Albina and Khan, Adil},
  journal={arXiv:1910.00373},
  year={2019}
}

@article{yin2020meta,
  title={{Meta-Learning for Few-Shot Natural Language Processing: A Survey}},
  author={Yin, Wenpeng},
  journal={arXiv: 2007.09604},
  year={2020}
}

@article{aumiller2022eur,
  title={{EUR-Lex-Sum: A Multi-and Cross-lingual Dataset for Long-form Summarization in the Legal Domain}},
  author={Aumiller, Dennis and Chouhan, Ashish and Gertz, Michael},
  journal={arXiv:2210.13448},
  year={2022}
}

@article{gunning1969fog,
  title={{The Fog Index After Twenty Years}},
  author={Gunning, Robert},
  journal={Journal of Business Communication},
  volume={6},
  number={2},
  pages={3--13},
  year={1969},
}

@article{mc1969smog,
  title={{SMOG Grading-a New Readability Formula}},
  author={Mc Laughlin, G Harry},
  journal={Journal of reading},
  volume={12},
  number={8},
  pages={639--646},
  year={1969},
}

\end{document}